\documentclass[11pt]{article}
\usepackage{graphicx,ifthen,color,algorithm,palatino}
\usepackage{amsmath,amsthm,amssymb,amsfonts,verbatim}
\usepackage{mathrsfs,setspace}
\newboolean{UnBlinded}
\newboolean{DoubleSpaced}
\setboolean{UnBlinded}{true}
\setboolean{DoubleSpaced}{false}
\def\Dsc{{\mathcal D}}
\def\phat{{\widehat p}}
\def\bdelta{\boldsymbol{\delta}}
\def\bh{\boldsymbol{h}}
\def\real{{\mathbb R}}
\def\bs{\boldsymbol{s}}
\def\Hsc{{\mathcal H}}
\def\bell{\boldsymbol{\ell}}
\def\bX{\boldsymbol{X}}
\def\ptrue{p_{\mbox{\tiny true}}}

\def\Csc{{\mathcal C}}
\def\smhalf{{\textstyle{\frac{1}{2}}}}

\def\infint{\int_{-\infty}^{\infty}}

\def\bbeta{\boldsymbol{\beta}}
\def\bI{\boldsymbol{I}}

\def\bone{\boldsymbol{1}}

\def\bmu{\boldsymbol{\mu}}
\def\bSigma{\boldsymbol{\Sigma}}

\def\bu{\boldsymbol{u}}
\def\simind{\stackrel{{\tiny \mbox{ind.}}}{\sim}}
\def\bZ{\boldsymbol{Z}}

\def\bC{\boldsymbol{C}}
\def\bc{\boldsymbol{c}}
\def\be{\boldsymbol{e}}
\def\bE{\boldsymbol{E}}
\def\utilde{{\widetilde u}}

\def\Bsc{{\mathcal B}}
\def\bib{\vskip12pt\par\noindent\hangindent=1 true cm\hangafter=1}
\def\Asc{{\mathcal A}}

\def\bv{\boldsymbol{v}}

\def\myand{\&\ }

\def\bbetabu{
\left[
\begin{array}{c}
\bbeta\\
\bu
\end{array}
\right]
}

\def\realnos{{\mathbb R}}

\begin{document}
\ifthenelse{\boolean{DoubleSpaced}}{\setstretch{1.5}}{}

\thispagestyle{empty}

\centerline{\LARGE\bf Density Estimation via Bayesian Inference Engines} 
\vskip7mm
\ifthenelse{\boolean{UnBlinded}}{
\centerline{\large\sc By M.P. Wand and J.C.F. Yu}
\vskip6mm
\centerline{\textit{University of Technology Sydney}}
\vskip6mm
\centerline{20th September, 2021}
}{\null}

\vskip6mm

We explain how effective automatic probability density function estimates 
can be constructed using contemporary Bayesian inference engines such 
as those based on  no-U-turn sampling and expectation 
propagation. Extensive simulation studies demonstrate that 
the proposed density estimates have excellent comparative
performance and scale well to very large sample sizes due to a binning strategy.
Moreover, the approach is fully Bayesian and all estimates are
accompanied by pointwise credible intervals. An accompanying package in the 
\textsf{R} language facilitates easy use of the new density estimates.


\section{Introduction}\label{sec:intro}

Bayesian inference engines have become established as an important 
paradigm for inference in arbitrarily large and complex
graphical models. Software platforms such as \textsf{Infer.NET} 
(Minka \textit{et al.}, 2018) and \textsf{Stan} (Carpenter \textit{et al.}, 2017)
are instances of such Bayesian inference engines. They deliver
approximate Bayesian inference, with varying degrees of inferential
accuracy, by calling upon contemporary approaches such as expectation propagation,
Hamiltonian Monte Carlo and variational approximation.
The purpose of this short article is to show that effective and scalable 
probability density function estimation, or density estimation for short, can be 
achieved using Bayesian inference engines. We provide easy access 
for users of the \textsf{R} statistical computing environment 
(\textsf{R} Core Team, 2018) via a package named \textsf{densEstBayes} 
\ifthenelse{\boolean{UnBlinded}}{(Wand, 2021)}{\null}.

Even though density estimators such as the histogram have had
a presence in statistics and data analysis for most of its
history, automatic density estimation started as a major area of 
research in the early 1980s when computing power aided its 
feasibility. Practical methodology, usually involving 
kernel density estimation with a data-driven bandwidth choice,
such as Rudemo (1982), Bowman (1984) and Sheather \myand Jones (1991) 
was accompanied by deep theoretical analysis such as Hall \myand Marron (1987). 
Several other proposals ensued, many of which are summarized in Chapter
3 of Wand \myand Jones (1995). A more recent proposal of this
general type is due to Botev, Grotowski \myand Kroese (2010), in which
kernel density estimation is combined with diffusion theory
to yield an advanced plug-in type bandwidth selector. A simulation study 
given there demonstrates superior practical performance compared with earlier 
proposals.

In a separate literature, starting mainly in the early 1990s, practical 
methodology for inference in Bayesian graphical models emerged
as a major area of activity. The most prominent approach is Markov
chain Monte Carlo which aims to produce samples from the posterior
density functions of hidden nodes (parameters and
latent variables) in a graphical model.
By the mid-1990s the \textsf{BUGS} Bayesian inference engine
(e.g.\ Lunn \textit{et al.}, 2009) had emerged and, for the first
time, data analysts could perform approximate Bayesian inference
for an arbitrarily complicated Bayesian graphical model by doing 
little more than specifying the model and inputting the data.
The last 25 years has seen various refinements of this 
paradigm. An interesting review of the state-of-affairs in
the mid-2000s is provided by Murphy (2007). Since that time
two new major Bayesian inference engines have emerged:
\textsf{Infer.NET} (Minka \textit{et al.}, 2018) and \textsf{Stan} 
(Carpenter \textit{et al.}, 2017). The former of these is distinguished 
by the fact that its main approaches to approximate Bayesian
inference are deterministic, rather than based on Monte Carlo
sampling, with expectation propagation (e.g.\ Minka, 2001) 
and mean field variational Bayes (e.g.\ Wainwright \myand Jordan, 2008)
being the underlying principles called upon.
The \textsf{Stan} Bayesian inference engine uses Hamiltonian
Monte Carlo and a variant known as the no-U-turn sampler
(Hoffman \myand Gelman, 2014) to obtain samples from
the posterior density functions of hidden nodes.
In an area with close ties to density estimation: 
nonparametric regression and various extensions,
Luts \textit{et al.} (2018) and Harezlak, Ruppert \myand Wand (2018)
provide several illustrations of approximate Bayesian inference
via \textsf{Infer.NET} and \textsf{Stan} respectively.

The more fundamental problem of automatic probability density
function estimation via Bayesian inference engines is the 
focus here. The crux of our approach is to express the 
density estimation problem as a Poisson nonparametric
regression problem. This involves replacement of the
original data by bin counts on a fine equally-spaced
grid (Eilers \myand Marx, 1996) as detailed in Section
\ref{sec:PNRjustif}. Bayesian Poisson nonparametric regression 
using mixed model representations of low-rank smoothing splines (e.g.\ Ruppert,
Wand \myand Carroll, 2009) can be expressed as
a Bayesian graphical model and is easy to feed into
a Bayesian inference engine. 
The conversion of the density estimation problem to 
that of fitting a Poisson nonparametric regression
model also has the advantage of scaling well to 
massive sample sizes since the only cost
for large sample sizes is the binning step. Once the
input data have been binned the remaining operations 
are unaffected by sample size. In Section \ref{sec:evaluation}
we report the results of a simulation study that 
demonstrates Bayesian inference engine density
estimation to be very accurate in comparison 
with existing methods. 

Various Bayesian approaches to density estimation have been proposed 
over the past few decades and articles on the topic
number in the dozens. Broad themes include use of continuous-time
stochastic processes (e.g. Leonard, 1978; Lenk, 1988) and nonparametric Bayes 
discrete-time stochastic process structures 
(e.g. Escobar \myand West, 1995; Petrone, 1999). Unlike the 
Poisson nonparametric regression/low-rank smoothing spline
model considered here, see (\ref{eq:PoissMixMod}) in Section \ref{sec:approach},
the majority of these approaches are not amenable to immediate
implementation in a Bayesian inference engine.

A Bayesian inference engine estimator at any
particular abscissa has a corresponding variability measure
-- nominally in the form of a 95\% credible interval.
This entails the option of adding a variability
band around the plotted density estimate which has the
advantage of providing a visualization of the sample variability.
Interpretation of variability bands requires caution since 
they are based on \emph{pointwise} credible intervals.
The problem of obtaining \emph{simultaneous} credible interval bands,
in the spirit of Sun \myand Loader (1994), is not explored here.
In Section \ref{sec:inferAcc} we demonstrate that
the empirical coverages of the credible intervals
produced by Bayesian inference engine density
estimation is somewhat conservative but usually 
meets advertized coverage levels. The density estimation
literature contains numerous proposals for the 
construction of confidence intervals
(e.g.\ Hall \myand Titterington, 1988; Chen, 1996; 
Gin\'e \myand Nickl, 2010) for frequentist inference
concerning density function values. The methodology
is generally of a high technical level, with
delicate asymptotic arguments and practical implementation
hindered by \emph{ad hoc} smoothing parameter choice.
In contrast, Bayesian inference engine density estimation
provides credible intervals in a simple and natural way.

Full details of our approach and some examples are presented in 
Section \ref{sec:approach}. In Section \ref{sec:evaluation} we report 
the results of simulation studies concerned with evaluation
of our proposal. We describe the \textsf{densEstBayes} \textsf{R} 
package in Section \ref{sec:Rpackage}. Conclusions are drawn in 
Section \ref{sec:conclusions}.

\section{Approach}\label{sec:approach}

We start by describing our Bayesian inference engine
approach to density estimation in generic form.
There are various choices to be made such as 
the actual Bayesian inference engine to use and
auxiliary parameters such as the number of 
spline basis functions and Bayesian model
hyperparameters. These choices are discussed in 
Sections \ref{sec:choiceBIE} and \ref{sec:choiceAuxPar}. 
A key feature of the approach is conversion of the density 
estimation problem to a Poisson nonparametric regression
problem. Its justification is given in Section \ref{sec:PNRjustif}.

The following notation is needed to describe the approach. 
A random variable $v$ has an Inverse Gamma distribution
with shape parameter $\kappa$ and scale parameter $\lambda$
if and only if its density function is 
$$p(v)=1/\{\lambda^{\kappa}\Gamma(\kappa)\}\,v^{-\kappa-1}\exp(-v/\lambda),\quad v>0,$$
and $p(v)=0$ for $v\le0$. The statement $v_i\simind\Dsc_i$, $1\le i\le n$,
means each of the random variables $v_i$ are independent with
distribution $\Dsc_i$.

Given a univariate random sample $x_1,\ldots,x_n$, the generic approach 
to obtaining an estimate of the sample's probability density function is:

\begin{enumerate}
\item Linearly transform the $x_i$, $1\le i\le n$, to the unit interval.
\item Replace the $x_i$, $1\le i\le n$, by bin counts on a fine equally-spaced
grid of size $M$ over the unit interval. Let $(g_{\ell},c_{\ell})$, $1\le \ell\le M$,
denote the grid point/grid count pairs.
The choice of $M$ is discussed in Section \ref{sec:choiceAuxPar}.
\item Fit the Bayesian mixed model-based penalized spline model:
\begin{equation}
\begin{array}{l}
c_{\ell}|\beta_0,\beta_1,u_1,\ldots,u_K\simind\mbox{Poisson}\left\{\exp\left(\beta_0+\beta_1\,g_{\ell}
+{\displaystyle\sum_{k=1}^K}u_kz_k(g_{\ell})\right)\right\},\\[3ex]
\beta_0,\beta_1\simind N(0,\sigma_{\beta}^2),\quad
u_1,\ldots,u_K|\sigma^2\simind N(0,\sigma^2),  
\\[2ex]
\sigma^2|a\sim\mbox{Inverse-Gamma}(\smhalf,1/a),\quad a\sim\mbox{Inverse-Gamma}(\smhalf,1/s_{\sigma}^2)
\end{array}
\label{eq:PoissMixMod}
\end{equation}
via some Bayesian inference engine. Its choice is discussed in Section \ref{sec:choiceBIE}.
Choice of the spline basis $\{z_k(\cdot):1\le k\le K\}$ and hyperparameters $\sigma_{\beta},s_{\sigma}>0$
is covered in Section \ref{sec:choiceAuxPar}.
\item For any $x\in[0,1]$, the density estimate of the transformed data is
\begin{equation}
\phat(x)=C^{-1}\left\{\mbox{posterior mean of}\ 
\exp\left(\beta_0+\beta_1\,x+\sum_{k=1}^K u_kz_k(x)\right)\right\}
\label{eq:phat}
\end{equation}
and $C$ is chosen to ensure that $\int_0^1\phat(x')\,dx'=1$.
Pointwise credible intervals to accompany the $\phat(x)$, $0\le x\le 1$, are readily
available from Step 3. Details are given in Section \ref{sec:credInt}.
\item Linearly transform the density estimate and corresponding credible intervals to the
original data units.
\end{enumerate}
Note that 
$$\sigma^2|a\sim\mbox{Inverse-Gamma}(\smhalf,1/a),\quad a\sim\mbox{Inverse-Gamma}(\smhalf,1/s_{\sigma}^2)$$
is equivalent to the standard deviation parameter $\sigma$ having a Half Cauchy prior
density function with scale parameter $s_{\sigma}$:
$$p(\sigma)=2/[\pi s_{\sigma}\{1+(\sigma/s_{\sigma})^2\}],\quad \sigma>0.$$
The use of the auxiliary variable $a$ in (\ref{eq:PoissMixMod}) aids 
the construction of approximate Bayesian inference schemes such
as those discussed in Sections \ref{sec:EP} and \ref{sec:SMFVB}.

The density estimate produced by steps 1.-5. takes the form of an exponentiated cubic spline,
where the coefficients are subject to a roughness penalty. The essence of this general approach
goes back, at least, to Boneva, Kendall \myand Stefanov (1971). Several articles, such as
Wahba (1975) and Good \myand Gaskins (1980), have built on this general paradigm. 
The class of density estimates presented in this section is of the same ilk, but uses
low-rank smoothing splines and takes advantage of the Bayesian inference engine revolution.

\subsection{Justification for Use of Poisson Nonparametric Regression}\label{sec:PNRjustif}

Conversion of the density estimation problem to a Poisson nonparametric 
regression via binning over a fine grid is a relatively old trick,
and is explained and used in Section 8 of Eilers \myand Marx (1996)
for a version of penalized spline-based nonparametric regression.
The justification hinges upon an equivalence between Poisson and Multinomial
maximum likelihood estimators as explained in Section 13.4.4 of 
Bishop, Fienberg \myand Holland (2007).

\subsection{Choice of Bayesian Inference Engine}\label{sec:choiceBIE}

Potentially, the only difficult step in Bayesian Poisson nonparametric density estimation
is fitting the Bayesian model (\ref{eq:PoissMixMod}). Established Bayesian
inference engines such \textsf{BUGS}, \textsf{Infer.NET}, \textsf{JAGS} (Plummer, 2003)
and \textsf{Stan} essentially remove this difficulty. At the time
of this writing the main costs are computing time and the occasional
need for chain diagnostics. Refinements of these packages and improved 
future Bayesian inference engines will continue to make the fitting of 
(\ref{eq:PoissMixMod}) faster and more routine.
Another option for fitting (\ref{eq:PoissMixMod}) is self-implementation of 
one of the very many approximate Bayesian inference schemes in the literature. 
For example, the ``stepping out'' slice sampling strategy of Neal (2003) has 
a particularly simple implementation and, if programmed in a low-level language, 
can be reasonably fast compared with the general purpose Markov chain 
Monte Carlo schemes used by established Bayesian inference engines.

In our exploration and demonstration of the efficacy of density estimation
via Bayesian inference engines we settled on four approaches, and these feature 
in the numerical evaluations given in Section \ref{sec:evaluation}. We now
provide some details on each of these four approaches.

\subsubsection{Expectation Propagation}\label{sec:EP}

Expectation propagation is a class of deterministic approximations of the joint posterior density
function of the parameters in a graphical model, based on notions such as Bethe free
energy and expectation constraints. A theoretical framework for expectation propagation
is provided by Heskes \textit{et al.} (2005), where the problem is expressed in Lagrangian 
form. However, the underlying optimization problem is challenging due to its non-convex 
saddle point nature. The most common strategy for obtaining practical solutions is to use  
iterative \emph{message passing} on an appropriate \emph{factor graph} with so-called 
\emph{damping} adjustments. Expectation propagation message passing updates correspond 
to Kullback-Leibler projections onto particular exponential family density functions 
and, intuitively, are iterative moment-matching operations.  Kim \myand Wand (2018) 
provide the algorithmic details of expectation propagation for various generalized, linear 
and mixed models.

\begin{figure}[!ht]
\centering
{\includegraphics[width=0.95\textwidth]{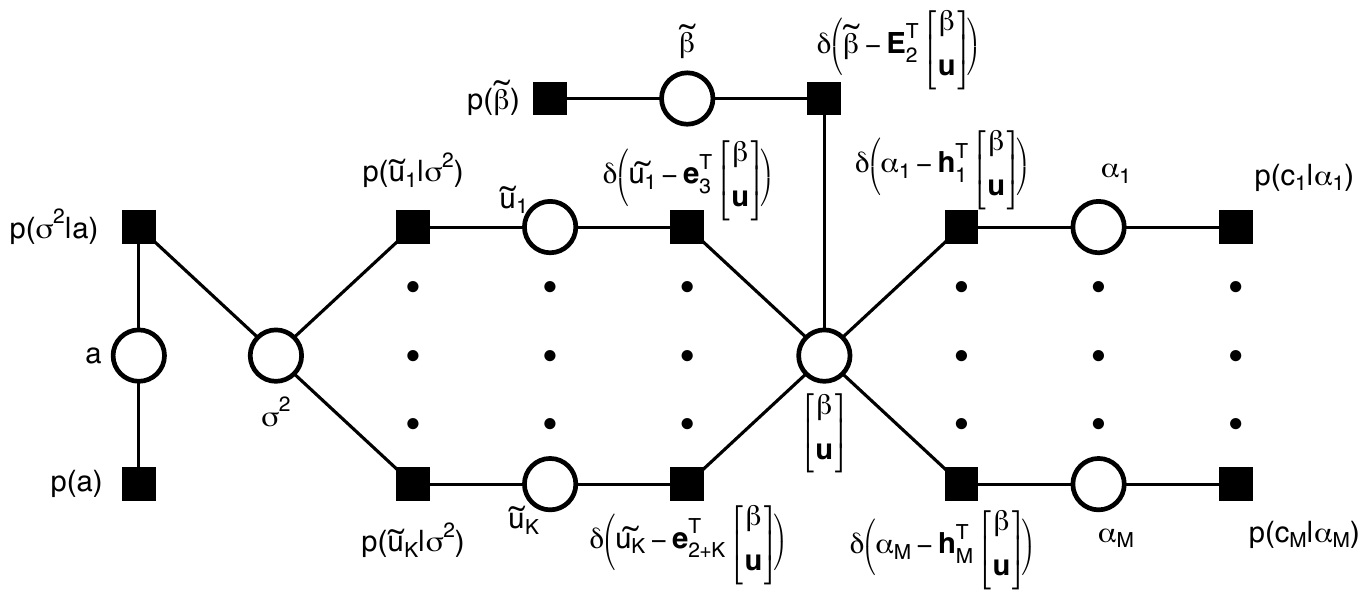}}
\caption{\it Derived variable factor graph corresponding to 
the representation of $p(\bc,\bbeta,\bu,\sigma^2,a)$ given by
(\ref{eq:pureRanEffFullFac}).}
\label{fig:densEstBayesEPfacGraph} 
\end{figure}

Model (\ref{eq:PoissMixMod}) is a special case of the  models treated in Kim \myand Wand (2018).
Let $\bbeta\equiv(\beta_0,\beta_1)$ and $\bu\equiv(u_1,\ldots,u_K)$ be the coefficient 
vectors and $\bc\equiv(c_1,\ldots,c_M)$ be the vector of bin counts.
The factor graph for expectation propagation approximation of the posterior
density functions
$$p(\bbeta,\bu|\bc)\quad\mbox{and}\quad p(\sigma^2|\bc),$$
is formed by noting the following algebraic truism:
\begin{equation}
{\setlength\arraycolsep{1pt}
\begin{array}{l}
p(\bc,\bbeta,\bu,\sigma^2,a)
=p(a)\,p(\sigma^2|\,a)\,
\left\{{\displaystyle\prod_{k=1}^K}\,\displaystyle{\infint}
p(\utilde_k|\,\sigma^2)\,\delta\left(\utilde_k-\be_{k+2}^T\bbetabu\right)
\,d\utilde_k \right\}\\[1ex]
\qquad\times
\left\{{\displaystyle\int_{\realnos^2}}
p({\widetilde\bbeta})\,
\bdelta\left({\widetilde\bbeta}-\bE_2^T\bbetabu\right)\,
d{\widetilde\bbeta}\right\}
\left\{{\displaystyle\prod_{\ell=1}^M}\,\displaystyle{\infint} 
p(c_{\ell}|\,\alpha_i)\delta\left(\alpha_\ell-\bh_{\ell}^T\bbetabu\right)\,d\alpha_{\ell}\right\}.
\end{array}
}
\label{eq:pureRanEffFullFac}
\end{equation}
Here $\delta$ denotes the univariate Dirac delta function, 
$\bdelta$ denotes the bivariate Dirac delta function, 
$\be_j$ is the $(2+K)\times1$ vector
with $1$ in the $j$th entry and all other entries equal to zero, $\bE_2$
is the $(2+K)\times2$ matrix with $\bI_2$ in first two rows and all other entries
equal to zero and $\bh_{\ell}\equiv(1,g_{\ell},z_1(g_{\ell}),\ldots,z_K(g_{\ell}))$.
Figure \ref{fig:densEstBayesEPfacGraph} is a factor graph representation
of (\ref{eq:pureRanEffFullFac}). The factors are shown as solid rectangles and 
stochastic variables as circles, with edges joining stochastic variables to the 
factors that include them. The integral signs in (\ref{eq:pureRanEffFullFac}) 
are ignored and Kim \myand Wand (2018) use the phrase \emph{derived variable} 
factor graph to make this distinction from regular factor graphs, with the 
$\alpha_{\ell}=\bh_{\ell}^T[\bbeta^T\ \bu^T]^T$ being examples of derived variables.

Bayesian density estimation via expectation propagation proceeds by 
updating messages passed  between each of the neighboring nodes 
on the Figure \ref{fig:densEstBayesEPfacGraph} and iteration until 
convergence. Full details are in Kim \myand Wand (2018). The message 
updates required evaluation of versions of the following non-analytic 
integral functions:
\begin{equation}
{\setlength\arraycolsep{2pt}
\begin{array}{rcl}
\Asc(p,q,r,s,t,u)&=&\displaystyle{
\infint\frac{x^p\,\exp(qx-rx^2)\,dx}{(x^2+sx+t)^u}},\\[2ex]
\Bsc(p,q,r,s,t,u)&=&\displaystyle{
\infint\frac{x^p\,\exp\{qx-re^x-se^x/(t+e^x)\}\,dx}{(t+e^x)^u}}\\[2ex]
\mbox{and}\qquad\Csc(p,q,r)&=&\displaystyle{\infint x^p\,\exp(qx-rx^2-e^x)\,dx}
\end{array}
}
\label{eq:ABCintegrals}
\end{equation}
with various restrictions on the parameters as detailed in Section 2.1 of Kim \myand Wand (2018).
Inversion of the function $\log-\mbox{digamma}$, where 
$\mbox{digamma}(x)\equiv\frac{d}{dx}\log\{\Gamma(x)\}$,
is also required. All other calculations are algebraic. 

\subsubsection{No U-Turn Sampling}\label{sec:NUTS}

No U-turn sampling, due to Hoffman \myand Gelman (2014), is a Markov chain Monte Carlo scheme 
that fine tunes Hamiltonian Monte Carlo sampling. Hamiltonian Monte Carlo, also known
as hybrid Monte Carlo, dates back to the mid-1980s statistical physics literature. Its
naming, and introduction to the mainstream statistics literature, is due to Neal (2011).
As explained there, Hamiltonian Monte Carlo is a version of Metropolis-Hastings-based
Markov chain Monte Carlo. The traditional approach to producing Metropolis-Hastings
proposal distributions involves random walks. As explained in Neal (2011), 
the posterior distribution space can be explored more efficiently when 
Metropolis-Hastings proposal distributions are instead produced using 
the principles of Hamiltonian dynamics. Full details are provided by this 
landmark article.

The essence of no U-turn sampling is adaptive choice of parameters that are inherent to 
Hamiltonian Monte Carlo, such as step size
and number of steps, via the introduction of slice variables. No U-turn sampling is the default 
and preferred algorithm in the \textsf{Stan} Bayesian inference engine for obtaining samples from 
the posterior distributions of hidden nodes in a graphical model. In recent years 
no U-turn sampling has established itself as a durable and high-quality Markov chain Monte
Carlo scheme. Almost all of the Bayesian semiparametric regression examples in Harezlak
\textit{et al.} (2018) use no U-turn sampling. Its availability within
the \textsf{R} package \textsf{rstan} means that (\ref{eq:PoissMixMod}) can be 
embedded within the \textsf{R} computing environment via just a few lines of code.

\subsubsection{Semiparametric Mean Field Variational Bayes}\label{sec:SMFVB}

Mean field variational Bayes aims to achieve approximate Bayesian inference
for (\ref{eq:PoissMixMod}) via a product density restriction approximations such as 
\begin{equation}
p(\bbeta,\bu,\sigma^2,a|\bc)\approx q(\bbeta,\bu)q(\sigma^2)q(a).
\label{eq:meanfield}
\end{equation}
and choosing the $q$-density functions to minimise the Kullback-Leibler
divergence of the right-hand side of (\ref{eq:meanfield}) from the
left-hand side. However, the form of the optimal $q$-density 
of the coefficients does not admit a closed
form. A practical remedy is the pre-specification
$$q(\bbeta,\bu)\ \ \mbox{is a}\ \  N(\bmu_{q(\bbeta,\bu)},\bSigma_{q(\bbeta,\bu)})
\ \ \mbox{density function}
$$
for some mean vector $\bmu_{q(\bbeta,\bu)}$ and covariance matrix $\bSigma_{q(\bbeta,\bu)}$,
followed by Kullback-Leibler minimization subject to this restriction.
This augmentation of mean field variational Bayes has various names
such as \emph{fixed-form variational Bayes} and \emph{non-conjugate variational message passing}.
Rohde \myand Wand (2016) make a case for the term \emph{semiparametric mean field variational Bayes},
and we use that label here. Model (\ref{eq:PoissMixMod}) is a special case of
the Poisson additive mixed models treated in Section 3.1 of Luts \myand Wand (2015).
The Poisson response and $r=1$ special case of Algorithm 1 in  Luts \myand Wand (2015)
leads to fast approximate Bayesian density estimation.

\subsubsection{Slice Sampling}

For random variables $s_1\in\real$, $s_2>0$, as well as random vectors $\bs_3$ and $\bs_4$,
of the same dimension, let 
$$x|s_1,s_2,\bs_3,\bs_4\sim\Hsc(s_1,s_2,\bs_3,\bs_4)$$
denote that the random variable $x$, conditional on 
$(s_1,s_2,\bs_3,\bs_4)$ has density function
\begin{equation}
p(x|s_1,s_2,\bs_3,\bs_4)\propto\exp
\left\{s_1x-x^2/(2s_2)-\bone^T\exp(x\bs_3+\bs_4)\right\},
\quad -\infty<x<\infty,
\label{eq:HscDens}
\end{equation}
where $\bone$ denotes a vector of ones having the same number of rows as
$\bs_3$ and $\bs_4$ and $\exp(x\bs_3+\bs_4)$ is evaluated element-wise.
Then scalar Gibbs sampling for (\ref{eq:PoissMixMod})
is such that draws are required from either density functions of the form
(\ref{eq:HscDens}) or Inverse Gamma density functions. The latter is 
trivial and the former is relatively easy if one uses the ``stepping out''
slice sampling approach of Neal (2003). Let $\bC\equiv[\bX\ \bZ]$ and define 
$\bC_j$ to be the $j$th column of $\bC$ and $\bC_{-j}$ to be the matrix $\bC$ 
with its $j$th column removed.
Similarly, define
$$
\left[
\begin{array}{c}
\bbeta\\
\bu
\end{array}
\right]_{-j}
\quad\mbox{to be} 
\quad
\left[
\begin{array}{c}
\bbeta\\
\bu
\end{array}
\right]
\quad\mbox{with its $j$th entry removed.}
$$
If $G$ denotes the total number of samples, including the warm-up, then
a suitable slice sampling within Gibbs sampling scheme is (after e.g.\ 
setting an initial value for $(\sigma^2)^{\,[0]}$):

\begin{minipage}[t]{120mm}
\begin{itemize}
\item[] For $g=1,\ldots,G$:
\begin{itemize}
\item[] $\bv\longleftarrow [\sigma_{\bbeta}^2\bone_2^T,
(\sigma^2)^{\,[g-1]}\bone_{K}^T]^T$
\item[] For $j=1,\ldots,2+K$:
\begin{itemize}
\item[] 
$\left[
\begin{array}{c}
\bbeta\\
\bu
\end{array}
\right]_j^{[g]}\sim \Hsc\left((\bC^T\bc)_j,\bv_j,\bC_j,
\bC_{-j}\left[
\begin{array}{c}
\bbeta\\
\bu
\end{array}
\right]_{-j}^{[g-1]}\right)$ 
\end{itemize}
\item[] $a^{[g]}\sim \mbox{Inverse-Gamma}\,
\Big(1,1/(\sigma^2)^{\,[g-1]}+1/s_{\sigma}^2\Big)$,
\item[]$(\sigma^2)^{[g]}\sim \mbox{Inverse-Gamma}
\Big(\smhalf(K+1),
\smhalf\Vert\bu^{[g]}\Vert^2+1/a^{[g]}\Big)$.
\end{itemize}
\end{itemize}
\end{minipage}

\vskip2mm
\noindent
where $\bc=(c_1,\ldots,c_M)$ is the vector of bin counts.

After omission of the warm-up samples, the $R$ retained samples 
$$\left[
\begin{array}{c}
\bbeta\\
\bu
\end{array}
\right]^{[g]},\quad 
(\sigma^2)^{[g]},\quad 1\le g\le R,$$
can be used for construction of a Bayesian density estimate and
pointwise credible intervals. In the Section \ref{sec:examples} examples
and Section \ref{sec:evaluation} simulation studies we used
a warm-up of size $100$ and $R=1,000$ retained samples.

\subsection{Choice of Auxiliary Parameters}\label{sec:choiceAuxPar}

Full specification of the Bayesian inference engine-based density
estimator requires choice of the basis functions, hyperparameters
and binning grid size. For most density functions that arise in
applications the choices of these auxiliary parameters have very 
little effect on the estimate. We provide good default settings here.
If the density function has intricate features and the sample size 
is very large to the extent that these features can be estimated
reasonably then some adjustment to these defaults may be required.

For the spline basis functions $\{z_k(\cdot):1\le k\le K\}$ we use 
cubic canonical O'Sullivan splines as described in Section 4
of Wand \myand Ormerod (2008). The default number of basis
functions is $K=50$. The number of grid points used for 
binning is defaulted to $M=401$ which is in keeping with 
Table 1 of Hall \myand Wand (1996). Linear binning
(e.g.\ Hall \myand Wand, 1996) is used, followed by rounding
to the nearest integer, to get the bin counts $c_{\ell}$ for
use in the Poisson nonparametric regression model.
The default hyperparameter values are $\sigma_{\beta}=s_{\sigma}=1,000$,
assuming the transformation of the input data to the unit
interval has taken place, corresponding to approximate noninformativity.

For the Monte Carlo-based approaches we ran a pilot study to test
the effect of the warm-up and retained sample sizes on density
estimation accuracy. For the no-U-turn sampler we found that
a warm-up of length $1,000$ with $1,000$ retained samples
was adequate without significant degradation of accuracy.
For the slice sampling a warm-up of length $100$, followed by
$1,000$ retained samples, was found to be adequate.
For expectation propagation and semiparametric mean field variational
Bayes the default stopping criterion is the relative change
in $E_q(1/\sigma^2)$ falling below $10^{-5}$.

\subsection{Pointwise Credible Interval Construction}\label{sec:credInt}

Pointwise credible intervals are a simple by-product of the 
Bayesian inference engine output. Suppose that $x_0\in[0,1]$ 
is a typical abscissae of interest. In the case of no-U-turn 
sampling the samples from the posterior density functions
of the $\beta_j$ and $u_k$ can be used to form a sample corresponding 
to the $\phat(x_0)$ according to the form given by (\ref{eq:phat}).
For $0<\alpha<1$, an approximate $100(1-\alpha)$\% credible interval for $p(x_0)$ has upper
and lower limits corresponding to the $\alpha/2$ and $(2-\alpha)/2$ sample
quantiles of this sample. 

For expectation propagation and semiparametric mean field variational 
Bayes we instead have $\bmu_{q(\bbeta,\bu)}$ and $\bSigma_{q(\bbeta,\bu)}$
as Bayesian inference engine outputs. If we let 
$$\bell(x_0)\equiv[1\ x_0\ z_1(x_0)\ \cdots\ z_K(x_0)]^T$$
be the vector of basis function evaluations at $x_0$ then, with $\Phi$ denoting
the $N(0,1)$ cumulative distribution function, 
$$\bell(x_0)^T\bmu_{q(\bbeta,\bu)}\pm\Phi^{-1}\big((2-\alpha)/2\big)
\sqrt{\bell(x_0)^T\bSigma_{q(\bbeta,\bu)}\bell(x_0)}$$
is an approximate $100(1-\alpha)$\% credible interval for the linear form
$$\beta_0\,x_0+\beta_1\,x_0+\sum_{k=1}^K u_kz_k(x_0).$$
Simple manipulations then lead to an approximate $100(1-\alpha)$\% credible interval for the $p(x_0)$.

\subsection{Pre-processing Options}\label{sec:preproc}

If the input data are strongly skewed or contain gross outliers then 
some pre-processing may be worthwhile. The fourth example of
the upcoming Figure \ref{fig:densEstBayesExamps} applies a logarithmic
transform to the input data. Bayesian inference engine 
density estimation is applied to these transformed data.
The estimate is back-transformed for graphical display.

\subsection{Examples}\label{sec:examples}

\begin{figure}[!ht]
\centering
{\includegraphics[width=0.87\textwidth]{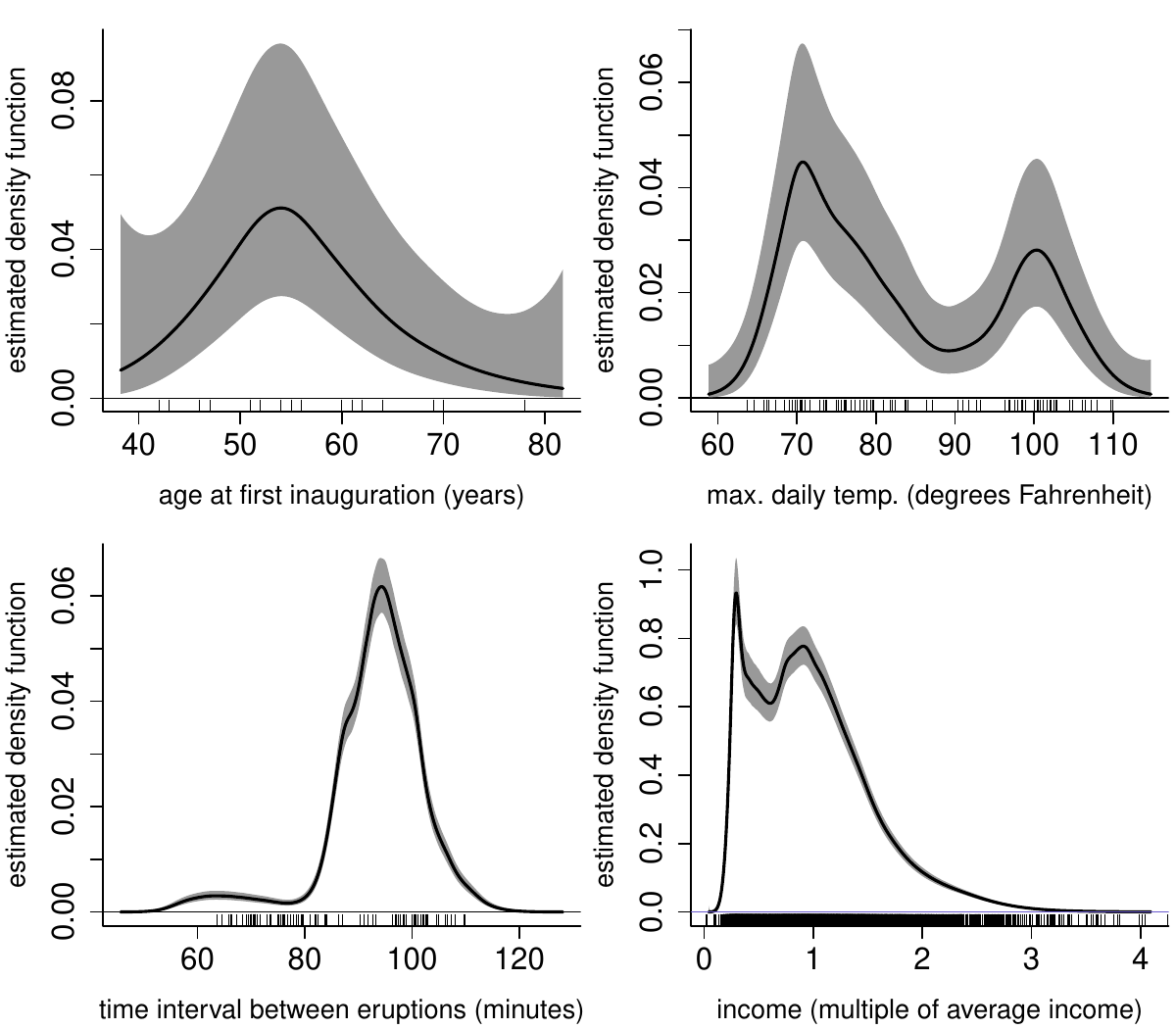}}
\caption{\it 
Examples of Bayesian inference engine density estimation,
with semiparametric mean field variational Bayes used in each case. The tick marks
at the base of each plot show the data and shaded regions
correspond to pointwise 95\% credible intervals. Top left: 
the data are the ages (years) at first inauguration of the 28 
U.S. presidents whom have held office during 1900--2021. Top right:
the data are maximum daily temperature (degrees Fahrenheit) 
in Melbourne, Australia, for the 101 days that followed
a very hot day, defined to be 95 degrees Fahrenheit or higher,
during 1981--1990. Bottom left: the data are time intervals (minutes) 
between all 3,507 adjacent pairs of eruptions of the Old Faithful Geyser 
during the 2011. Bottom right: the data are incomes of 7,201 
United Kingdom citizens for the year 1975. The data have been divided 
by average income.}
\label{fig:densEstBayesExamps} 
\end{figure}

Figure \ref{fig:densEstBayesExamps} provides four examples of 
Bayesian inference density estimation for the following univariate
data sets:

\begin{itemize}
\item[] ages in years at first inauguration of the 29 presidents of the United States 
of America who have held office during 1900--2021; 
\item[] maximum daily temperature in degrees Fahrenheit in Melbourne, Australia, for the 
101 days that followed a very hot day, defined to be 95 degrees Fahrenheit or higher,
during 1981--1990;
\item[] time intervals in minutes between all 3,507 adjacent pairs of eruptions of the 
Old Faithful Geyser in Yellowstone National Park, U.S.A, during 2011, obtained
from The Geyser Observation and Study Association web-site (\texttt{www.geyserstudy.org});
\item[] incomes of 7,201 United Kingdom citizens for the year 1975, divided by average income. 
The source of these data is the Economic and Social Research Council Data Archive at the 
University of Essex, United Kingdom.
\end{itemize}

Since the data sets increase in size from the top left panel to the bottom 
right panel the 95\% credible intervals become narrower. The last three density
estimates having interesting bimodal structure. For the maximum daily 
temperatures in Melbourne the bimodality is explained by the southerly buster 
phenomenon, which often produces a dramatic temperature drop after a very hot day
in southern Australia.

\section{Evaluation}\label{sec:evaluation}
 
The new density estimation strategies described in Section \ref{sec:approach}
add to a large field of existing automatic density estimators. 
We now investigate how they compare in terms of accuracy
and computing time.

\subsection{Density Estimation Accuracy}\label{sec:densEstAccur}

\begin{figure}[!ht]
\centering
{\includegraphics[width=\textwidth]{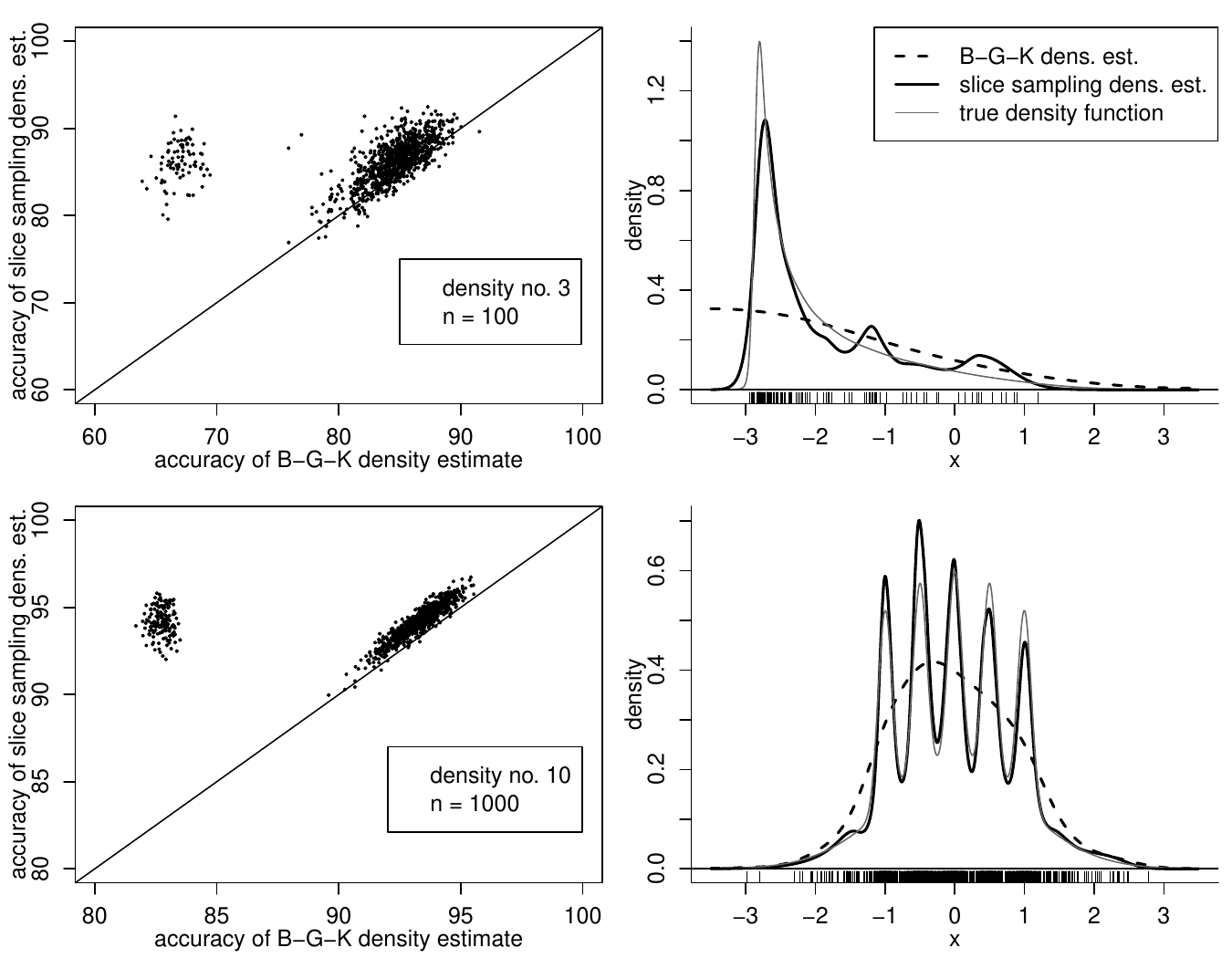}}
\caption{\it Simulation study results that demonstrate improvements
of Bayesian inference engine density estimation over an existing
approach. Top row: Estimation of Marron-Wand density number 3 with
a sample size of $n=100$. The top left panel scatterplot shows 
accuracy values of estimates based on slice sampling versus those
based on the Botev-Grotowski-Kroese approach. The 1:1 line is also 
plotted. The upper left cluster represents a significant practical 
improvement of the Bayesian inference engine approach and contains 
8.1\% of the replications. The top right shows example density estimates 
corresponding to this cluster. Bottom row: Analogous to the bottom 
row but for Marron-Wand density number 10 with $n=1,000$. 
The upper left cluster contains 14.9\% of the replications.
}
\label{fig:BGKvsSLICEclus} 
\end{figure}

We ran a large simulation study involving 3 sample sizes,
10 true density functions and 6 automatic density function
estimators. The sample sizes are $n\in\{100,1000,10000\}$,
the true density functions are density numbers 1--10 in
Table 1 of Marron \myand Wand (1992). The density estimation
methods are (in order of development):
\begin{itemize}
\item[] kernel density estimation with bandwidth chosen according to least squares
cross-validation (Rudemo, 1982; Bowman, 1984),
\item[] kernel density estimation with bandwidth chosen according a direct plug-in
strategy as described in Section 3.6.1 of Wand \myand Jones (1995),
\item[] the diffusion kernel density estimator of Botev, Grotowski \myand Kroese (2010),
\item[] and the four types of Bayesian inference engine density estimators described
in Section \ref{sec:approach} --- involving each of expectation
propagation, no-U-turn sampling, semiparametric mean field variational Bayes 
and slice sampling.
\end{itemize}
Estimation accuracy of a generic density estimate $\phat$ was measured using the 
accuracy score
$$\mbox{accuracy}(\phat)=100\left(1-\smhalf\int_{-\infty}^{\infty}\Big|\phat(x)-\ptrue(x)\Big|\,dx\right)$$
which uses the fact that the $L_1$ error $\int_{-\infty}^{\infty}|\phat(x)-\ptrue(x)|\,dx$
is a scale-free number between $0$ and $2$ and linearly transforms this error 
measure to an accuracy percentage. The simulation study was run over $1,000$ replications.
For each pair of methods, the accuracy paired difference samples were analyzed using visual 
inspections of side-by-side box plots and Wilcoxon confidence intervals.
The main findings were as follows:
\begin{itemize}
\item The four Bayesian inference engine approaches were such that there was very little practical differences
between them in terms of accuracy. Further investigations have revealed that the choice of Bayesian 
inference engine has a negligible effect on the density estimate in terms of visual appearance.
\item The diffusion kernel density estimator usually dominated the other kernel density estimates
in terms of accuracy. For some settings such as Marron-Wand density number 5 there
were pronounced practical improvements of the diffusion kernel density estimator compared
with ordinary kernel density estimation with least squares cross-validation bandwidth choice.
\item In 29 out of the 30 settings both the no-U-turn-based and slice sampling-based density
estimator exhibited a statistically significant better accuracy than the diffusion kernel density 
estimator. Although for most of these settings the practical improvement was negligible,
there were some practical improvements in a few settings. Figure \ref{fig:BGKvsSLICEclus}
and its accompanying discussion describes these improvements.
\end{itemize}

Figure \ref{fig:BGKvsSLICEclus} shows some of the practical advantages of Bayesian inference engine
density estimation over a state-of-the art approach. The upper panels are for sample sizes of $n=100$
and estimation of the third Marron-Wand density function, which is strongly skewed. The scatterplot
in the upper left panel of Figure \ref{fig:BGKvsSLICEclus} shows that, whilst most of the time
the two approaches have very similar accuracies, about 8\% of the replication are such that 
the diffusion kernel density estimate suffers from a pronounced drop in accuracy --- corresponding
to the cluster above the 1:1 line.  The top right panel shows typical estimates from
this cluster, with the diffusion kernel density estimate over-smoothing. The lower panels
of Figure \ref{fig:BGKvsSLICEclus} tell a similar story for samples of size $n=1,000$ 
and estimation of the tenth Marron-Wand density function, which is claw-shaped.

\subsection{Computing Time}

We kept track of the computing times in the simulation study described in the
previous subsection. The default auxiliary parameter values described in 
Section \ref{sec:choiceAuxPar} were used. The simulations were run on 
a \textsf{MacBook Air} laptop computer with a 2.2 gigahertz processor and 8 gigabytes
of random access memory. Of course, the computing times are impacted by choices 
such as warm-up length and hardware specifications. Nonetheless, the
results presented here give an idea of computing time using typical early 2020s
personal hardware, as well as comparative performance. 

Table \ref{tab:timeResults} provides the $10$th, $50$th and $90$th quantiles
of the computing times in seconds for each approach. Semiparametric mean
field variational Bayes is the fastest by far and usually returns an estimate in
less than half a second. However, as explained in Section \ref{sec:numericIssues},
this speed has to be counterbalanced against occasional convergence failure problems.
Despite our implementation in a low-level language, expectation propagation can be 
quite slow due to the large number of numerical integrations that it requires.
The no-U-turn and slice sampling approaches typically take about 10 to 15 seconds.
The computing time for slice sampling is less variable, with a 90th percentile
of 15 seconds compared with 24 seconds for no-U-turn sampling.

\begin{table}[!ht]
\centering
\begin{tabular}{lllll}
\hline\\[-1.5ex]
                   &$\quad$ & 10th percentile & 50th percentile & 90th percentile\\[0.5ex]
\hline\\[-1.2ex]
expect. propagation&   & 19         &  31           &    78   \\
no-U-turn sampling &   & \ \ 6.9    &  12           &    23  \\
semipar. MFVB      &   & \ \ 0.076  & \ \ 0.21     & \ \ 1.2 \\
slice sampling     &   & \ \!13    & 13            & \!\ 15  \\[0.2ex]
\hline
\end{tabular}
\caption{\it Percentiles for the computing times in seconds for each
of the Bayesian inference engine approaches used in the simulation study 
described in Section \ref{sec:densEstAccur}.
}
\label{tab:timeResults} 
\end{table}

\subsection{Bayesian Inferential Accuracy}\label{sec:inferAcc}

\begin{figure}[!ht]
\centering
{\includegraphics[width=0.87\textwidth]{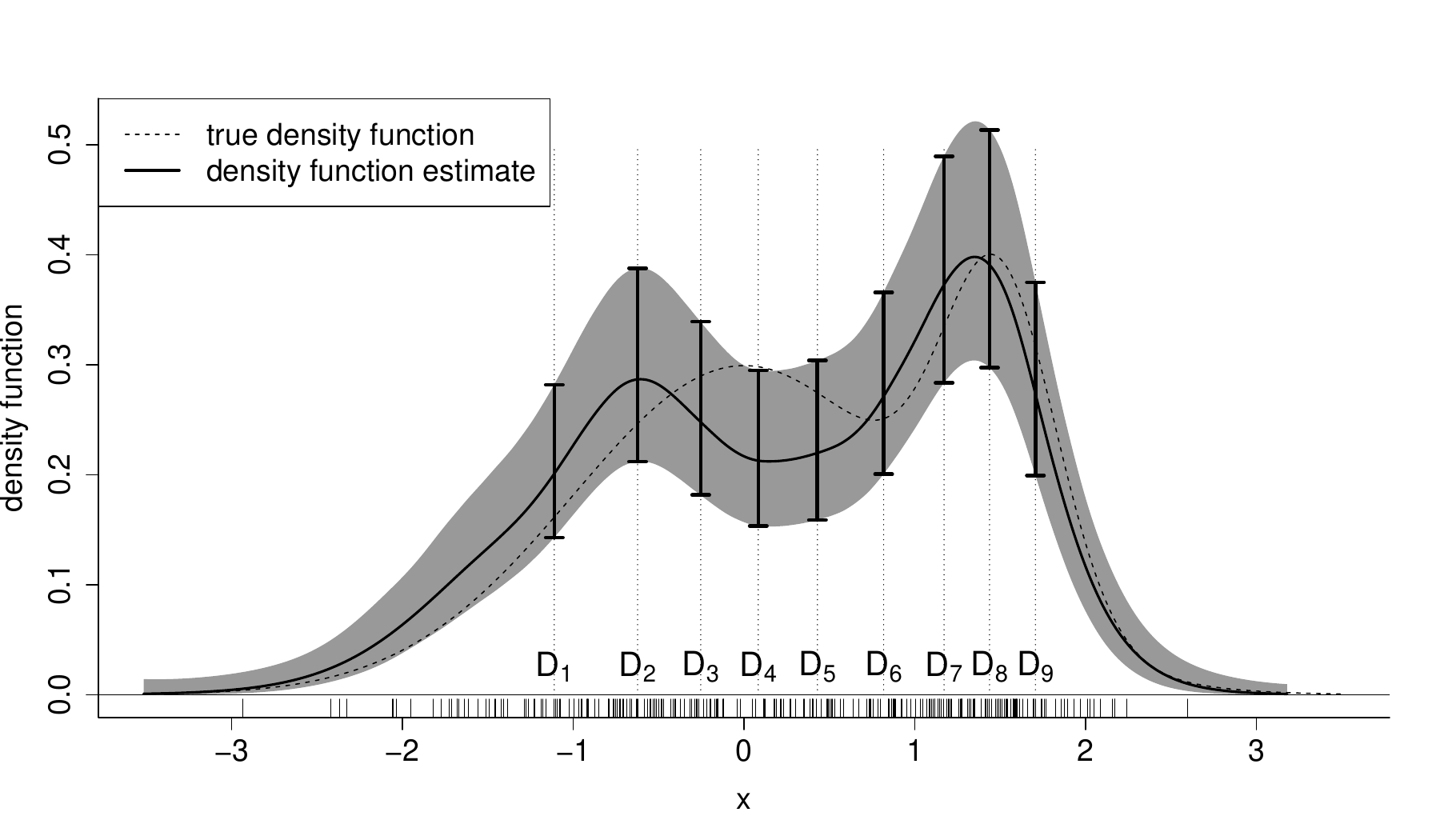}}
\caption{\it The density function given at (\ref{eq:MWdens8}) and the 
semiparametric mean field variational Bayes estimate based on a 
random sample of size $n=250$. The vertical line segments and notches
indicate 95\% credible intervals for $\ptrue(D_j)$ at each of
the deciles $D_j$, $1\le j\le 9$.}
\label{fig:decilesDens8} 
\end{figure}

In a second simulation study we investigated the degree to which
the pointwise credible sets produced by the proposed Bayesian
density estimates meet their advertized coverage levels.
Figure \ref{fig:decilesDens8} shows the semiparametric mean
field variational Bayes-based density estimate from on a sample 
size of $n=1,000$ generated from the eighth density function 
in Table 1 of Marron \myand Wand (1992):
\begin{equation}
\ptrue(x)=3\Big[\exp\big(-x^2/2\big)
+\exp\big\{-9\,(x-\textstyle{\frac{3}{2}})^2/2\big\}\Big]\Big/(4\sqrt{2\pi}).
\label{eq:MWdens8}
\end{equation}
We focused on inference for each of $\ptrue(D_1),\ldots,\ptrue(D_9)$
where $D_1,\ldots,D_9$ are the population deciles of $\ptrue$.
Figure \ref{fig:decilesDens8} shows the locations of the $D_j$
along with 95\% credible intervals for each $\ptrue(D_j)$ using
an $n=250$ Bayesian density estimate via the semiparametric
mean field variational Bayes approach. In this example eight of the
nine credible intervals cover the true density function value.
The exception is $\ptrue(D_4)$, which is not quite covered by 
its 95\% credible interval.

To assess coverage accuracy, for sample sizes $n=100$, $n=1,000$ and $n=10,000$, 
we generated $1,000$ random samples and obtained density estimates using each of the
three Bayesian inference engines described in Section \ref{sec:choiceBIE}.
Table \ref{tab:credIntSimul} shows the empirical coverage percentages.

\begin{table}[!ht]
\begin{center}
\begin{tabular}{llllllllllll}
&        &$D_1$&$D_2$&$D_3$&$D_4$&$D_5$&$D_6$&$D_7$&$D_8$&$D_9$\\     
 \hline		

$n=100$  & EP   & 98.4 & 98.7& 98.8& 98.8& 98.0& 93.2& 98.3& 95.8& 92.6 \\ 
         & NUTS & 97.7 & 98.4& 99.0& 99.2& 97.5& 93.9& 98.8& 92.7& 95.5 \\ 
         & SMFVB& 97.5 & 98.0& 99.0& 99.2& 97.6& 90.6& 98.6& 90.2& 95.5 \\ 
         & slice& 97.6 & 98.3& 99.0& 99.5& 97.5& 93.9& 98.7& 92.2& 95.1 \\ 
\hline	
$n=1,000$ & EP   & 98.3& 98.8& 98.9& 98.9& 98.7& 96.1& 98.9& 98.4& 96.9 \\
         & NUTS & 98.4& 98.3& 98.7& 98.8& 98.9& 95.7& 98.7& 96.5& 97.3 \\
         & SMFVB& 98.5& 98.3& 98.5& 98.9& 98.9& 94.7& 99.0& 96.7& 97.7 \\
         & slice& 98.5& 98.6& 98.8& 98.6& 98.9& 95.3& 98.8& 97.1& 97.8 \\
\hline

$n=10,000$ & EP   & 98.3& 98.6& 98.4& 98.8& 98.2& 97.5& 98.3& 98.1& 98.4 \\ 
           & NUTS & 97.1& 98.4& 98.0& 98.5& 97.7& 97.9& 98.5& 98.8& 97.9 \\ 
           & SMFVB& 97.3& 98.3& 97.9& 98.9& 97.7& 97.7& 98.6& 98.9& 98.2 \\ 
           & slice& 97.2& 98.4& 97.7& 98.6& 98.0& 97.7& 98.5& 98.6& 98.1 \\ 
\hline
\end{tabular}
\end{center}
\caption{\it Empirical coverage percentages for the Bayesian inference problem
conveyed by Figure \ref{fig:decilesDens8} involving 95\% credible intervals
at each of the deciles of the density function given at (\ref{eq:MWdens8}).
The methods are no U-turn sampling (NUTS), expectation propagation (EP),
semiparametric mean field variational Bayes (SMFVB) and slice sampling (slice).
The number of replications was $1,000$.}
\label{tab:credIntSimul} 
\end{table}

We see from Table \ref{tab:credIntSimul} that, in almost every case, the 
empirical coverage level exceeds the 95\% advertized coverage level.
This indicates that inference based on the proposed Bayesian inference 
engine density estimators is honest in that it delivers on what it
promises. However, with an average empirical coverage of 97.6\%
it is apparent that the estimators over-deliver compared with 
their 95\% advertizement. The conclusion from this limited
study concerning Bayesian accuracy of the proposed Bayesian inference 
engine density estimators is that they are honest although they err on 
the side of conservatism.

\subsection{Numerical Issues}\label{sec:numericIssues}

The semiparametric mean field variational Bayes approach involves 
fixed point iteration to find the minimum Kullback-Leibler divergence 
$\bmu_{q(\bbeta,\bu)}$ and $\bSigma_{q(\bbeta,\bu)}$ parameters.
For the default of 50 spline basis functions there are $1,430$
free parameters in this search. Often convergence is successful
and rapid. However, despite efforts to obtain good starting values,
the semiparametric mean field variational Bayes approach failed
to converge for 13.6\% of the $30,000$ data sets in the Section 
\ref{sec:densEstAccur} simulation study. Therefore, further
numerical analytic research is required to make semiparametric mean field 
variational Bayes more practical. Expectation propagation
converged in almost all $30,000$ data sets but sometimes could
be quite slow taking as long as several minutes on a 2020s laptop. 
Despite being a deterministic alternative to Monte Carlo sampling, 
the integrals that arise when expectation propagation is applied to 
model (\ref{eq:PoissMixMod}) require quadrature and are also 
quite numerous. This leads to expectation propagation often 
being considerably slower than the Monte Carlo-based approaches.
Further research is warranted for speeding up expectation propagation
to acceptable levels.

\section{Accompanying \textsf{R} Package}\label{sec:Rpackage}

An \textsf{R} package  that 
accompanies this article is available on the Comprehensive \textsf{R} 
Archive Network (\texttt{https://www.R-project.org})
and named \textsf{densEstBayes} 
\ifthenelse{\boolean{UnBlinded}}{(Wand, 2021)}{\null}.
Once installed, the following few commands illustrate its default use:

\begin{verbatim}
library(densEstBayes) ; x <- rnorm(1000) 
dest <- densEstBayes(x) ; plot(dest) ; rug(x)
\end{verbatim}

\noindent
The other arguments of the \texttt{densEstBayes()} function,
named \texttt{method} and \texttt{control}, respectively allow for 
different Bayesian inference engines to be specified and auxiliary
parameters to be controlled. In the version of \textsf{densEstBayes}
that is available at the time of this writing the Bayesian inference
engines are Hamiltonian Monte Carlo, no-U-turn sampling, semiparametric
mean field variational Bayes and slice sampling. The last of these
is the default method due to it achieving a good balance in terms of
accuracy performance, numerical reliability and speed.

The \textsf{densEstBayes} package is accompanied by vignette which provides
fuller details on its use. The vignette PDF file is opened via the command
\texttt{densEstBayesVignette()}. 

\section{Conclusions}\label{sec:conclusions}

We have demonstrated that Bayesian inference engines based
on the Poisson nonparametric regression and mixed model-based
penalized splines offer a competitive class of density
estimators, and sometimes lead to noticeable improvements
in accuracy compared with state-of-the-art approaches.
Moreover, Bayesian inference engines are accompanied by
principled variability bands which enhance graphical display.
Our recommended default Bayesian inference engine takes
about $10$--$15$ seconds to compute on early 2020s laptop computers,
which is a reasonable price to pay given its attractive
attributes. Future Bayesian inference engines and hardware
enhancements offer the prospect of further improvement.

\section*{Acknowledgments}

\ifthenelse{\boolean{UnBlinded}}{
We are grateful for assistance from Eman Alfaifi.
This research was supported by Australian Research Council
Discovery Project DP140100441. 
}{\null}

\section*{References}

\bib
Bishop, Y.M.M., Fienberg, S.E. \myand Holland, P.W. (2007).
\textit{Discrete Multivariate Analysis: Theory and Practice.}
New York: Springer.

\bib
Boneva, L.I., Kendall, D. \myand Stefanov, I. (1971).
Spline transformations: three new diagnostic aids for the
statistical data-analyst  (with discussion). 
\textit{Journal of the Royal Statistical Society, Series B}, 
{\bf 33}, 1--71.

\bib
Botev, Z.I., Grotowski, J.F. \myand D.P. Kroese (2010).
Kernel density estimation via diffusion.
\textit{The Annals of Statistics}, {\bf 38}, 2916--2957.

\bib
Bowman, A.W. (1984). An alternative method of cross-validation
for the smoothing of density estimates.
\textit{Biometrika}, {\bf 71}, 353--360.

\bib
Carpenter, B., Gelman, A., Hoffman, M.D., Lee, D., Goodrich, B., Betancourt, M., 
Brubaker, M., Guo, J., Li, P. \myand Riddell, A. (2017). 
Stan: A probabilistic programming language. 
\textit{Journal of Statistical Software}, {\bf 76}, Issue 1, 1--32.

\bib
Chen, S.X. (1996). Empirical likelihood confidence intervals for
nonparametric density estimation. {\it Biometrika}, {\bf 83}, 
329--341.

\bib
Escobar, M.D. \myand West, M. (1995). Bayesian density estimation
and inference using mixtures. \textit{Journal of the American
Statistical Association},
{\bf 90}, 577--588.

\bib
Eilers, P.H.C. \myand Marx, B.D. (1996).
Flexible smoothing with B-splines and penalties
(with discussion).
{\it Statistical Science}, {\bf 11}, 89--121.

\bib
Gin\'e, E. \myand Nickl, R. (2010). 
Confidence bands in density estimation.
{\it The Annals of Statistics}, {\bf 38}, 1122-1170.

\bib
Good, I.J. \myand Gaskins, R.A. (1980).
Density estimation and bump-hunting by the penalized likelihood
method exemplified by scattering and meteorite data
(with discussion).
\textit{Journal of the American Statistical Association},
{\bf 75}, 42--73.

\bib
Hall, P. \myand Marron, J.S. (1987). Extent to which least-squares
cross-validation minimised integrated square error in nonparametric
density estimation. \textit{Probability Theory and Related Fields},
{\bf 74}, 567--581.

\bib
Hall, P. \myand Titterington, D.M. (1988). 
On confidence bands in nonparametric density estimation
and regression. {\it Journal of Multivariate Analysis},
{\bf 27}, 228--254.

\bib
Hall, P. \myand Wand, M.P. (1996). On the accuracy of binned kernel
density estimators. {\it Journal of Multivariate Analysis}, {\bf 56},
165--184.

\bib
Harezlak, J., Ruppert, D. \myand Wand, M.P. (2018).
{\it Semiparametric Regression with R}.
New York: Springer.  

\bib
Heskes, T., Opper, M., Wiegerinck, W., Winther, O. \myand Zoeter, O. (2005).
\textit{Journal of Statistical Mechanics: Theory and Experiment, P11015},
1--24. 

\bib
Hoffman, M.D. \myand Gelman, A. (2014). 
The no-U-turn sampler: adaptively setting path
lengths in Hamiltonian Monte Carlo. \textit{Journal of Machine Learning Research},
{\bf 15}, 1593--1623.

\bib
Kim, A.S.I. \myand Wand, M.P. (2018).
On expectation propagation for generalised, linear and mixed models. 
\textit{Australian and New Zealand Journal of Statistics},
{\bf 60}, 75--102.

\bib
Lenk, P.J. (1988). The logistic normal distribution for Bayesian, nonparametric,
predictive densities. \textit{Journal of the American Statistical Association},
{\bf 83}, 509--516.

\bib
Leonard, T. (1978). Density estimation, stochastic processes, and prior information.
\textit{Journal of the Royal Statistical Society, Series B}, {\bf 40}, 113--146.

\bib
Lunn, D., Spiegelhalter, D., Thomas, A. \myand Best, N. (2009).
The \textsf{BUGS} project: evolution, critique and future
directions. \textit{Statistics in Medicine}, {\bf 28}, 3049--3067.

\bib
Luts, J. \myand Wand, M.P. (2015).
Variational inference for count response semiparametric
regression.
\textit{Bayesian Analysis}, {\bf 10}, 991--1023.

\bib
Luts, J., Wang, S.S.J., Ormerod, J.T. \myand Wand, M.P. (2018).
Semiparametric regression analysis via \textsf{Infer.NET}.
\textit{Journal of Statistical Software}, {\bf 87}, 
Issue 2, 1--37.

\bib
Marron, J. S. \myand Wand, M. P. (1992). Exact mean integrated squared error.
{\it The Annals of Statistics}, {\bf 20}, 712--736 .

\bib
Minka, T.P. (2001).
Expectation propagation for approximate Bayesian inference.
In J.S. Breese \myand D. Koller (eds),
\textit{Proceedings of the Seventeenth Conference on Uncertainty in
Artificial Intelligence}, pp. 362--369.
Burlington, Massachusetts: Morgan Kaufmann.

\bib
Minka, T., Winn, J., Guiver, J., Zayov, Y., Fabian, D.
\myand Bronskill (2018). 
Infer.NET 0.3, Microsoft Research Cambridge.
\texttt{http://dotnet.github.io/infer}.

\bib
Murphy, K. (2007). Software for graphical models: a review.
\textit{International Society for Bayesian Analysis Bulletin},
{\bf 14}, 13--15.

\bib
Neal, R. (2003). Slice sampling (with discussion). 
\textit{The Annals of Statistics}, {\bf 31}, 705--767.

\bib
Neal, R. (2011). MCMC using Hamiltonian dynamics.
In S. Brooks, A. Gelman, G.L. Jones and X.-L. Meng (eds),
\textit{Handbook of Markov Chain Monte Carlo}, pp. 113--162.
Boca Raton, Florida: CRC Press.

\bib
Petrone, S. (1999). Bayesian density estimation using Bernstein
polynomials. \textit{Canadian Journal of Statistics}, {\bf 27},
105--120.

\bib
Plummer, M. (2003). \textsf{JAGS}: a program for analysis of Bayesian
graphical models using Gibbs sampling.
In K. Hornik, F. Leisch and A. Zeileis, editors,
\textit{Proceedings of the 3rd International Workshop on 
Distributed Statistical Computing}.

\bib
\textsf{R} Core Team (2018). 
\textsf{R}: A language and environment for statistical computing.
\textsf{R} Foundation for Statistical Computing.
Vienna, Austria. \texttt{https://www.R-project.org}

\bib
Rohde, D. \myand Wand, M.P. (2016).
Semiparametric mean field variational Bayes: General principles
and numerical issues. \textit{Journal of Machine Learning Research}, 
{\bf 17(172)}, 1--47.

\bib
Rudemo, M. (1982). Empirical choice of histograms and kernel density
estimators. \textit{Scandinavian Journal of Statistics}, {\bf 9},
65--78.

\bib
Ruppert, D., Wand, M.P. \myand Carroll, R.J. (2009).
Semiparametric regression during 2003-2007.
{\it Electronic Journal of Statistics}, {\bf 3}, 1193--1256.

\bib
Sheather, S.J. \myand Jones, M.C. (1991). A reliable data-based
bandwidth selection method for kernel density estimation.
\textit{Journal of the Royal Statistical Society, Series B},
{\bf 53}, 683--690.

\bib
Sun, J. \myand Loader, C.R. (1994).
Simultaneous confidence bands for linear regression and
smoothing. \textit{The Annals of Statistics}, {\bf 22}, 1328--1345.

\bib
Wainwright, M.J. \myand Jordan, M.I. (2008).
Graphical models, exponential families, and variational inference.
\textit{Foundations and Trends in Machine Learning}, {\bf 1}, 1--305.

\bib
Wahba, G. (1975). Interpolating spline methods for density
estimation I. Equi-spaced knots. \textit{The Annals of Statistics}, 
{\bf 3}, 30--48.

\bib
Wand, M. P. \myand Jones, M. C. (1995). {\it Kernel Smoothing}.
London: Chapman and Hall.

\bib
Wand, M.P. \myand Ormerod, J.T. (2008).
On semiparametric regression with O'Sullivan penalized splines.
{\it Australian and New Zealand Journal of Statistics},
{\bf 50}, 179--198.

\bib
Wand, M.P. (2021). \textsf{densEstBayes}:
Density estimation via Bayesian inference algorithms.
\textsf{R} package version 1.0. \texttt{http://cran.r-project.org}.

\end{document}